\documentclass[lettersize,journal]{IEEEtran}
\usepackage{amsmath,amsfonts}
\usepackage{algorithmic}
\usepackage{algorithm}
\usepackage{array}
\usepackage{textcomp}
\usepackage{stfloats}
\usepackage{url}
\usepackage{verbatim}
\usepackage{graphicx}
\usepackage{cite}
\usepackage{booktabs}
\usepackage{multirow, hhline}
\usepackage{subcaption}
\usepackage{flushend}
\usepackage{array}
\begin{document}

\title{Towards Less Constrained Macro-Neural Architecture Search}

\author{Vasco Lopes \qquad Luís A. Alexandre 
\thanks{
This work was supported by `FCT - Fundação para a Ciência e Tecnologia' through the research grant `2020.04588.BD', and partially supported by NOVA LINCS (UIDB/04516/2020) with the financial support of FCT, through national funds and by CENTRO-01-0247-FEDER-113023 - DeepNeuronic.
}
\thanks{V. Lopes and L. A. Alexandre are with NOVA Lincs, Universidade da Beira Interior (email: vasco.lopes@ubi.pt).}
}

\markboth{V. Lopes and L. A. Alexandre: Towards Less Constrained Macro-Neural Architecture Search}{IEEE Transactions on Neural Networks and Learning Systems}


\maketitle

\begin{abstract}
Networks found with Neural Architecture Search (NAS) achieve state-of-the-art performance in a variety of tasks, out-performing human-designed networks. However, most NAS methods heavily rely on human-defined assumptions that constrain the search: architecture's outer-skeletons, number of layers, parameter heuristics and search spaces. Additionally, common search spaces consist of repeatable modules (cells) instead of fully exploring the architecture's search space by designing entire architectures (macro-search). Imposing such constraints requires deep human expertise and restricts the search to pre-defined settings. In this paper, we propose LCMNAS, a method that pushes NAS to less constrained search spaces by performing macro-search without relying on pre-defined heuristics or bounded search spaces. LCMNAS introduces three components for the NAS pipeline: i) a method that leverages information about well-known architectures to autonomously generate complex search spaces based on Weighted Directed Graphs with hidden properties, ii) an evolutionary search strategy that generates complete architectures from scratch, and iii) a mixed-performance estimation approach that combines information about architectures at initialization stage and lower fidelity estimates to infer their trainability and capacity to model complex functions. We present experiments in 13 different data sets showing that LCMNAS is capable of generating both cell and macro-based architectures with minimal GPU computation and state-of-the-art results. More, we conduct extensive studies on the importance of different NAS components in both cell and macro-based settings. Code for reproducibility is public at \url{https://github.com/VascoLopes/LCMNAS}.
\end{abstract}


\section{Introduction}
\begin{figure}[!t]
    \centering
    \includegraphics[width=1\columnwidth]{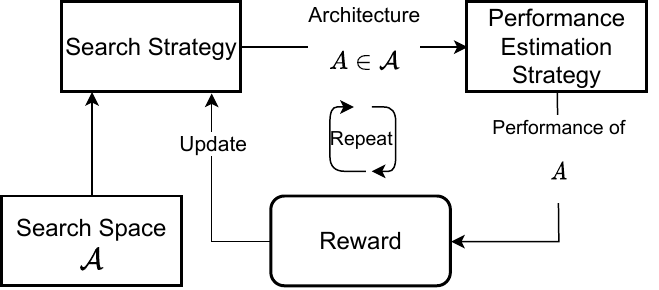}
    \caption{Generalist NAS flow. A controller generates an architecture $A$ from the space of possible architectures, $\mathcal{A}$, which is then evaluated, and its performance is used as reward to update the controller. In this paper, we propose methods for all components: search space, search strategy, performance estimation strategy. \label{fig:nasdiagram}}
\end{figure}

\IEEEPARstart{A}{dvances} in deep learning algorithms obtained remarkable progress in various problems, mainly due to the ingenuity and engineering efforts of human experts that exhaustively designed and engineered powerful architectures. Convolutional Neural Networks (CNNs) have been extensively applied with great success to different problems, obtaining unprecedented results \cite{lecun2015deep,schmidhuber2015deep,khan2020survey}. However, designing CNNs is a gruelling endeavour that heavily relies on human expertise. 
Thus, automating this process became logical \cite{hutter2019automated}. Neural Architecture Search (NAS) intends to automate architecture engineering and design \cite{elsken2019neural}. NAS methods have successfully been applied to image classification tasks \cite{cai2018proxylessnas,liu2018darts,zela2020understanding}, semantic segmentation \cite{liu2019auto,DBLP:conf/eccv/LiuDHGYX20}, object detection \cite{chen2019detnas} and image generation \cite{gong2019autogan,gao2020adversarialnas}, consistently achieving state-of-the-art results. Generally, NAS methods are composed of three components: i) the search space, which defines the pool of possible operations and thus, the type of networks that can be designed; ii) the search strategy, which is the approach used to explore the search space and generate architectures; and iii) the performance estimation strategy, which is how the generated architectures are evaluated during the search process. A general interaction between the different NAS components is presented in Fig. \ref{fig:nasdiagram}.
Architecture search is usually performed using either micro or macro-search. In micro-search, methods focus on creating cells or blocks that are replicated multiple times, whereas, in macro-search, NAS methods try to evolve entire network architectures. An increasing number of NAS methods have been proposed, where the focus has been targeted at optimizing the search to reduce the required computations, and at the same time, increase the performance of the generated architectures \cite{elsken2019neural,DBLP:journals/csur/RenXCHLCW21,liu2021survey}. 

Even though NAS methods perform well on designing CNNs, they still encounter several drawbacks: 
i) search spaces are heavily dependent on human-definitions, and are usually small with forced operations;
ii) the search is mostly cell-based, where NAS methods search for small cells that are later replicated in a human-defined outer-skeleton;
iii) architecture-related parameters, such as the number of layers, inner-layer parameters (e.g., the kernel size, output channels), the final architecture skeleton, fixed operations, and head and tail of the final architectures are usually defined by the authors;
and iv) the time required to perform macro-search (search entire networks) is still considerable.
By forcing rules and carefully designing search spaces, there is undoubtedly human biases introduced in the loop, which was found to often impact more the final result of the architectures than the search strategy itself, as it pushes NAS to constrained search spaces with very narrow accuracy ranges \cite{Yang2020NAS,wan2022on}, thus jeopardizing the generalization of the NAS methods, even to more simpler settings \cite{Dong2020NAS-Bench-201,DBLP:conf/uai/LiT19}. Moreover, the idea of heading to NAS to avoid needing deep knowledge regarding architecture design ends up being frustrated since a similar degree of human involvement is required as when designing isolated CNN architectures.

In this work, we propose \textbf{LCMNAS} - Towards \textbf{L}ess \textbf{C}onstrained \textbf{M}acro-\textbf{N}eural \textbf{A}rchitecture \textbf{S}earch, a NAS method that is capable of: i) autonomously generating complex search spaces by creating Weighted Directed Graphs with hidden properties (WDG) from existing architectures to leverage information that is the result of years of expertise, practice and trial-and-error; ii) performs macro-architecture search, without explicitly defined architecture's outer-skeletons, restrictions or heuristics; and iii) uses a mixed-performance strategy for estimating the efficiency of the generated architectures, that combines information about the architectures at initialization stage with information about their validation accuracy after a partial train on a partial data set. To validate the proposed method, we conduct extensive experiments in a macro-search setting, but also in micro-cell based search to allow further comparison with existing NAS methods. Extensive ablation studies show the importance of different NAS components in both micro and macro-search setting and discuss their usability and importance.

\begin{figure*}[!tbh]
\centering
\includegraphics[width=1\textwidth]{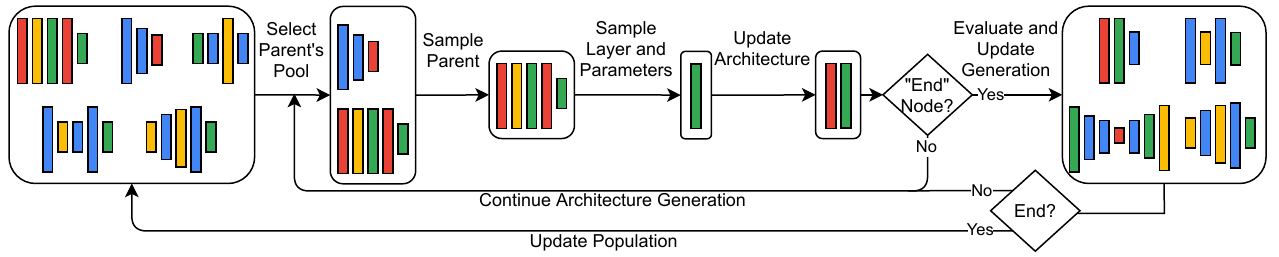} 
\caption{Illustration of one iteration of LCMNAS. Architectures are represented with varying width and bar's length to illustrate their diversity. The process shows the sequential design of sampling one layer and associated hyper-parameters for an architecture. Designing and evaluating architectures happens several times in a generation. The whole search ends when the evolution reaches a final generation.}
\label{fig:wrapfig}
\end{figure*}

The main contributions of this paper are summarized as follows:
\begin{itemize}
    \item We propose a search space design method that leverages information about existing CNNs to autonomously design complex search spaces.
    \item An evolutionary search strategy that goes beyond micro-search to macro-search. where entire architectures are designed, without forced architecture shape or structure, or well-engineered protocols. More, the proposed search strategy goes beyond solely designing architectures to also determining the hyper-parameters associated with each layer.
    \item A mixed-performance estimation mechanism that correlates untrained architecture scores with their final performance, and combines it with partial train to infer the architecture's trainability with minimal training. This allows for remarkable computation efficiency improvements.
    \sloppy\item Experiments demonstrate that the architectures found with LCMNAS achieve state-of-the-art results in both micro and macro-search settings, spanning across 13 different data sets.
    \item Extensive ablation studies show the applicability of different NAS components, e.g., zero-cost proxy estimation in both micro and macro-search settings, and evaluation of standard practices, such as transferring searched architectures to new data sets, and architecture diversity via ensembling. 
    \item Finally, we extract insights about architecture design based on the decisions made by LCMNAS that might inspire future research. 
\end{itemize}

\section{Related Work}
\textbf{Neural Architecture Search} was initially formulated as a Reinforcement Learning (RL) problem, where a controller was trained to sample more efficient architectures over time \cite{DBLP:journals/corr/ZophL16}. Albeit yielding excellent results, it required more than 60 years of GPU computation to search for an architecture. Follow-up work \cite{Zoph_2018}, proposed a cell-based search in a search space of 13 operations. To form entire architectures, generated cells were stacked according to pre-defined rules. Even thought it still required more than 2000 days of GPU computation, it prompted many proposals to focus on cell-based designs. Cell-based search spaces are generally represented as Directed Acyclic Graphs (DAGs), where nodes represent tensors and edges are operations \cite{elsken2019neural}. Different NAS strategies have been proposed to improve upon initial results, including novel RL strategies \cite{baker2016designing,DBLP:conf/cvpr/ZhongYWSL18,ENAS,howard2019searching} and evolutionary algorithms, where architectures are designed using evolution via cross-over and mutations \cite{real2019regularized,liu2018hierarchical,liu2021survey,xie2021benchenas,lopes2021guided,DBLP:conf/gecco/LopesSDA22}.

Recently, differentiable-based NAS attracted interest by designing cell-based architectures through relaxing a discrete search space and optimizing the search process using efficient gradient descent \cite{liu2018darts,xie2018snas,chen2019progressive,cai2018proxylessnas,Xu2020PC-DARTS:,zela2020understanding}. DARTS initially proposed the use of a bi-level gradient optimization to search for architectures and weights directly \cite{liu2018darts}. DARTS search space is popular, being used in the surrogate benchmark NAS-Bench-301 \cite{siems2020bench}. Common search spaces include tabular benchmarks, such as NAS-Bench-101 with 3 operations \cite{pmlr-v97-ying19a} and NAS-Bench-201 benchmark with 5 operations \cite{Dong2020NAS-Bench-201}, and the NASNet \cite{real2019regularized} with 8 operations. These search spaces are focused on the design of cell-based architectures.

Methods that perform macro-search are closely related to our work. NAS \cite{DBLP:journals/corr/ZophL16}, ENAS \cite{ENAS}, MetaQNN \cite{baker2016designing} and Net Transformations \cite{cai2018efficient} perform macro-search using RL strategies. Evolutionary algorithms have also been proposed \cite{elsken2018efficient,DBLP:conf/gecco/LuWBDDGB19,real2017large}, where architectures are designed through the use of mutations. Network Morphism has also been studied for macro-search, where morphisms are applied to initial architectures \cite{elsken2018efficient}. RandGrow does this based on random-search \cite{humacro}, and a follow-up work, Petridish, used gradient boost to further improve the method \cite{hu2019efficient}. Differently, EPNAS performs macro-search by Sequential Model-Based Optimization (SMBO) with weight sharing. However, the aforementioned macro-search NAS methods tend to strictly force human-defined rules, such as well-designed search spaces with very narrow accuracy ranges, architecture's outer-skeleton, their size (number of layers) and hyper-parameter heuristics. These forced rules have been shown to influence more the final accuracy of the architectures than the search itself \cite{DBLP:conf/uai/LiT19,Yang2020NAS,wan2022on}. NAS \cite{DBLP:journals/corr/ZophL16} and ENAS \cite{ENAS} also design additional filters for the best architectures by hand. From the proposed macro-search NAS methods, those that constrain the search less tend to be inefficient: LEMONADE required 80 GPU days of computation \cite{elsken2018efficient}, and large-scale evolution required 2750 GPU days \cite{real2017large}.

\textbf{Performance Estimation Strategy} reefers to the mechanism of evaluating the generated architectures and is usually the most costly component of a NAS method. Albeit training an architecture for a large number of epochs is the most reliable mechanism to rank it, this is extremely inefficient \cite{DBLP:journals/corr/ZophL16,Zoph_2018,DBLP:conf/eccv/LiuZNSHLFYHM18}. Several performance estimation strategies have been proposed to improve the search procedure. Weight sharing reduces the evaluation time by performing weight inheritance across generated architectures \cite{xie2018snas,ENAS,cai2018proxylessnas}. Low-fidelity estimates reduce the training time by training for fewer epochs or using subsets of the data, \cite{DBLP:journals/corr/abs-1807-06906}. Model-based predictors extrapolate the final validation accuracy of an architecture based on its encodings \cite{10.5555/3326943.3327130,Li_2020_CVPR,DBLP:conf/aaai/WhiteNS21}. Zero-cost proxies look at architectures at initialization stage, and calculate statistics that correlate with the architecture's final validation accuracy \cite{mellor2020neural,lopes2021epenas,DBLP:conf/iclr/ChenGW21}. White et. al. provide a comprehensive evaluation of 31 different performance estimation strategies \cite{white2021how}, showing that optimal results are obtained when combining multiple strategies.


To push NAS to less constrained settings, we go beyond micro-search and focus on designing methods that allow macro-search without forced settings and heuristics. For this, we propose a representation of complex search spaces as WDG, an evolutionary strategy that designs architectures from scratch, and an efficient mixed estimation-performance method that combines the evaluation of an architecture trainability through low fidelity estimates and its capability to model complex functions at initialization stage using a zero-proxy estimator.



\section{Proposed Method}
\label{sec:proposedmethod}

\subsection{Search Space}
\label{subsec:searchspace}
Search spaces are commonly defined as DAGs, where edges represent operations and nodes are tensors. The goal is to design cells that are repeated to form entire architectures, where the operations pool and architectures outer-skeletons are pre-defined by the users. However, this has been shown to force the search to very narrow accuracy ranges and to compromise the generalization beyond the evaluated data sets \cite{Yang2020NAS,DBLP:conf/uai/LiT19,wan2022on}, and at the same time, requires deep expertise to apply to new problems.

Instead of explicitly defining a search space, we propose a novel method to automatically design search spaces without requiring human-specified settings by leveraging information about existing CNNs, which are the result of years of expertise, practice and many trial-and-error experiments. For this, we extend the use of DAGs, and represent existing CNNs as Weighted Directed Graphs with hidden properties: $\mathbf{G(Q,E,H)}$, where $\mathbf{Q}$ is the set of nodes, $\mathbf{E}$ the set of edges and associated weights as state-transition probabilities, and $\mathbf{H}$ is the set of inner states of the nodes and associated probabilities.

To generate a WDG, $\mathbf{G(Q,E,H)}_i$, one can define a mapping from the input $x$ to an output $y$ using an architecture, $n_i$, as $w_{n_i}(x)$. This mapping is the result of feed-forwarding the input $x$ through a set of layers. By looking at the mapping, a summary of $n_i$ can be generated and used to produce the sets $\mathbf{Q}$ and $\mathbf{E}$. To obtain the state-transition probability between two layers, $P(l_i|l_j)$, we divide the frequency of the state-transition $C(l_i,l_j)$ by the sum of frequencies of state-transitions from $l_i$ to any other possible layer: $C(l_i,l_k), k=1,...,K$, where $K$ is the number of layers that appear after the layer $l_i$ in the summary. Thus, the weight of an edge $e(l_i,l_j) \in \mathbf{E}$ is given by: 
\begin{equation}
    e(l_i,l_j)=P(l_i|l_j)=\frac{C(l_i,l_j)}{\sum_{k=1}^{K}C(l_i,l_k)}
\end{equation}

The inner-states $h_i \in \mathbf{H}$ of a layer $l_i$, represent all possible parameters and associated values that appear in the CNN, e.g., for a convolutional layer, possible inner-states will be the output channels and kernel sizes and their associated probabilities. The inner-states of a layer $h \in \mathbf{H}$ are calculated similarly to the edges between nodes. For each possible component (e.g., kernel size), the probability associated with the value (e.g., kernel size $3\times3$) is calculated based on the frequency of that value divided by all possible values for that specific parameter. Note that a start and end node are added to all WDG to allow the search strategy to stop the search.

At the end, the search space, $\mathcal{A}$, is represented by all CNNs in the form of their WDGs and fitness for the given problem (see Section \ref{subsec:performance}), $\mathbf{(G(Q,E,H)}, f)$, such that $\mathcal{A} = \{(\mathbf{G(Q_1,E_1,H_1)}, f_1), ..., (\mathbf{G(Q_N,E_N,H_N)}, f_N)\}$, where $N$ is the number of CNNs used to create the initial search space.

By using WDGs to represent CNNs, the proposed method extracts information about the combination of layers and their associated parameters. This allows NAS methods to extract past information and move beyond solely designing layers and using heuristics to select their parameters, to a broader search, where the focus is on designing entire architectures by finding both the layers and associated parameters. The method is, thus, inferring human-expertise directly from the search space without requiring human intervention in the form of defined settings or parameters. Moreover, by combining the evaluation of CNNs for a specific problem in the form of the fitness $f$, into the WDGs, $\mathbf{G(Q,E,H)}$, the search strategy is capable of guiding the search more efficiently.

\begin{figure}[!tbh]
\centering
\includegraphics[width=0.5\columnwidth]{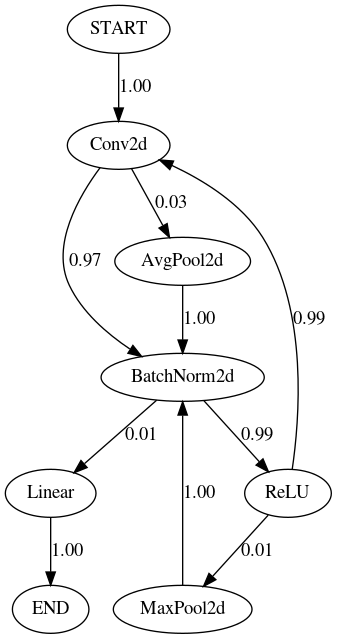} 
\caption{Weighted Directed Graph representation of DenseNet121. Nodes represent layers and edges represent state-transition probabilities between 2 nodes.\label{fig:lcmnasdensenet121}}
\end{figure}

In Fig. \ref{fig:lcmnasdensenet121}, the WDG of DenseNet121 is depicted, where layers are the nodes and connecting edges represent state-transition probabilities.

\subsection{Search Strategy} 
\label{subsec:searchstrategy}
We propose an evolutionary strategy that leverages information present in the WDGs, to generate architectures from scratch, without requiring heuristics or pre-defined schemes for the layer's hyper-parameter and architecture structures. Architectures are generated by sequentially sampling layers and associated hyper-parameters from the individuals present in $\mathcal{A}$. To generate a layer, $l_i$, a pool of parents, $D \subseteq \mathcal{A}$, is sampled based on the presence of the last sampled layer $l_{i-1}$ on all individuals from $\mathcal{A}$ ($l_{1}$ is the start node). An individual $j \in D$, is selected to be the parent of the layer $l_i$ by performing a ranked roulette wheel selection using the fitness of the individuals in $D$. This ensures that individuals in $D$ that have high fitness scores do not dominate the selection, thus promoting exploration of the search space. After an individual $j \in D$ is selected as parent, a layer $l_i$ is sampled using Fitness Proportionate Selection (FPS) by having as weights the state-transition probabilities present in $\mathbf{G(Q_j,E_j,H_j)}$, denoted as $P(l_t | l_{i-1}),~t = 1, ..., T$, where $T$ is the number of the nodes that have an input state-transition from $l_{i-1}$. Finally, the components of the layer sampled, e.g., output channels, are also sampled using FPS from the inner-state of the layer sampled $l_i$ from the sampled parent $j \in D$. The iterative process of sampling layers from different parents promotes diversity for the generated architecture. Furthermore, following the insights obtained by evolutionary NAS methods like large-scale evolution \cite{real2019regularized}, regarding the benefits of applying mutations, LCMNAS also employs a mutation process: when a convolutional layer is sampled, it has a fixed probability, $50\%$, of being mutated to a skip-connection. 

The evolutionary strategy is performed for $g$ generations, where at each generation, $p$ individuals are generated and the top $15\%$ architectures from the parents population are passed to the next generation through elitism. Fig. \ref{fig:wrapfig} illustrates the evolutionary process, where architectures are represented with varying width and bar lengths, and colors serve the purpose of differentiating between layers but do not represent specific types.



\subsection{Performance Estimation Mechanism}
\label{subsec:performance}
The greatest NAS bottleneck is the evaluation of the generated architectures, as fully training each one is extremely expensive, requiring thousands of GPU days of computation \cite{DBLP:journals/corr/ZophL16,Zoph_2018}. Accordingly, we propose a mixed-performance estimation approach that combines low fidelity estimates with zero-cost proxies to speed up the evaluation, while at the same time ensuring that the resulting score is a reliable approximation of the fully trained architecture ranking.

First, the proposed method evaluates the trainability of a generated architecture in a partial data set for a small number of epochs, $e$. Let the objective function that calculates the accuracy of the architecture $n_i$ on a small validation set, $d^{valid}$ be denoted by $\mathcal{O}(n_i,d^{valid})$. The validation accuracy is used as an indication if the architecture is capable of learning from the small number of examples shown, which ultimately can be used to distinguish between architectures that can be trained efficiently from those that cannot. The proposed method then looks at the capability of the untrained architecture at initialization stage to model complex functions, through Jacobian analysis \cite{mellor2020neural,lopes2021epenas}. For this, one can define a mapping from the input $\mathbf{x}_i \in \mathbb{R}^{D}$, through the network, $w(\mathbf{x}_i)$, where $\mathbf{x}_i$ represents an image that belongs to a batch $\mathbf{X}$, and $D$ is the input dimension. Then, this mapping can be computed by: $\mathbf{J}_i = \frac{\partial w(\mathbf{x}_i)}{\partial \mathbf{x}_{i}}$.

To evaluate how an architecture behaves for different images, we calculate the Jacobian $\mathbf{J}$ for different data points, $\mathbf{x}_i$, of the batch $\mathbf{X}$, $i \in 1, \cdots, I$:
\begin{equation}
\mathbf{J} = 
\begin{pmatrix}
\frac{\partial w(\mathbf{x}_1)}{\partial \mathbf{x}_{1}} & \frac{\partial w(\mathbf{x}_2)}{\partial \mathbf{x}_{2}} & \cdots & \frac{\partial w(\mathbf{x}_I)}{\partial \mathbf{x}_{I}} \\
\end{pmatrix}^{\top}
\label{eq:J}
\end{equation}

Then, the correlation matrix, $\mathbf{\Sigma}_J$, allows the analysis of how the architecture models the target function at different data points, where ideally, for different images, the architecture would have different $\mathbf{J}$ values, thus resulting in low correlated mappings. Then, we can use KL divergence to score an architecture based on the eigenvalues of its correlation matrix. Let $\sigma_{J, 1} \leq \dots \leq \sigma_{J, V}$ be the $V$ eigenvalues of $\mathbf{\Sigma}_J$. The untrained architecture can be scored as: 
\begin{equation}
s =  1\times10^4/\sum_{i=1}^{V} [ \log(\sigma_{J,i} + k) + (\sigma_{J,i} + k)^{-1}],
\end{equation}

where $k=1\times10^{-5}$.

The proposed mixed-performance estimation approach calculates the fitness, $f$, by combining the evaluation of the architecture's trainability, using $\mathcal{O}(n_i,d^{valid})$, and their capability of modeling complex functions, using $s$ by:
\begin{equation}
    f_{n_i} = (1-\lambda) \times \mathcal{O}(n_i,d^{valid}) + \lambda \times s
\end{equation}

where $\lambda$ serves the purpose of giving different weights to each component. When $\lambda=0$, the fitness of an architecture is only based on the trainability of the network by looking at the validation accuracy using a partial train on a partial data set, $\mathcal{O}(n_i,d^{valid})$. When $\lambda=1$, the architecture's capability of modeling complex functions at initialization stage is the only considered factor. In the conducted experiments (see Section \ref{sec:exps}) we evaluate the importance of $\lambda$.


\section{Designing Entire Architectures: Experiments and Results Analysis}
\label{sec:exps}
\subsection{Data Sets}
We directly search and evaluate the proposed method on three data sets: CIFAR-10 \cite{cifar10}, CIFAR-100 \cite{cifar10} and ImageNet16-120 \cite{Dong2020NAS-Bench-201}. CIFAR-10 and CIFAR-100 both contain 50K training images and 10k test images with $32\times32$ pixel sizes, having 10 and 100 classes, respectively. ImageNet16-120 is a down-sampled variant of ImageNet, where all images have $16\times16$ pixels and only considers the first 120 classes of ImageNet, resulting in 151.7K training images, 3K validation images and 3K test images. For searching purposes, we generated a partial data set for all the data sets. Here, 8\% and 2\% of the training set were randomly sampled to serve as partial training and partial validation sets. For training the final searched architectures, we use the original, entire splits.

\begin{table*}[!t]
\caption{Comparison of the proposed method against ResNet, the best architecture in NAS-Bench-201 benchmark and 3 types of random-search. The proposed method is evaluated using different $\lambda$ values for the proposed mixed-performance estimation strategy. The comparison is measured, when applicable, by the test error (\%), the inference time (ms), the search cost in GPU days and the number of parameters of the generated architecture (in millions).\label{tab:totalresults}}
\centering
\resizebox{\textwidth}{!}{%
\begin{tabular}{lc@{\hspace{0.5\tabcolsep}}c@{\hspace{0.5\tabcolsep}}c@{\hspace{0.5\tabcolsep}}cc@{\hspace{0.5\tabcolsep}}c@{\hspace{0.5\tabcolsep}}c@{\hspace{0.5\tabcolsep}}cc@{\hspace{0.5\tabcolsep}}c@{\hspace{0.5\tabcolsep}}c@{\hspace{0.5\tabcolsep}}c}
\toprule
\multicolumn{1}{c}{\multirow{2}*{Architecture}} & \multicolumn{4}{c}{CIFAR-10}      & \multicolumn{4}{c}{CIFAR-100}        & \multicolumn{4}{c}{ImageNet16-120}         \\ \cmidrule(l{2pt}r{2pt}){2-5} \cmidrule(l{2pt}r{2pt}){6-9} \cmidrule(l{2pt}r{2pt}){10-13}
 &
  \begin{tabular}[c]{@{}c@{}}Test Error\\ (\%)\end{tabular} &
  \begin{tabular}[c]{@{}c@{}}Inference \\ Time (ms)\end{tabular} &
  \begin{tabular}[c]{@{}c@{}}Search Cost\\ (GPU days)\end{tabular} &
  \begin{tabular}[c]{@{}c@{}}Params\\ (M)\end{tabular} &
  \begin{tabular}[c]{@{}c@{}}Test Error\\ (\%)\end{tabular} &
  \begin{tabular}[c]{@{}c@{}}Inference \\ Time (ms)\end{tabular} &
  \begin{tabular}[c]{@{}c@{}}Search Cost\\ (GPU days)\end{tabular} &
  \begin{tabular}[c]{@{}c@{}}Params\\ (M)\end{tabular} &
  \begin{tabular}[c]{@{}c@{}}Test Error\\ (\%)\end{tabular} &
  \begin{tabular}[c]{@{}c@{}}Inference\\ Time (ms)\end{tabular} &
  \begin{tabular}[c]{@{}c@{}}Search Cost\\ (GPU days)\end{tabular} &
  \begin{tabular}[c]{@{}c@{}}Params\\ (M)\end{tabular}  
   \\ \midrule 
ResNet  \cite{he2016deep}         & 6.43 & 0.24 $\pm$ 0.19         & {-}    & 1.7   & 29.14 & 0.21 $\pm$ 0.13 & {-}    & 1.7  & 56.37 & 0.10 $\pm$ 0.19    & {-}            & 1.7   \\
\begin{tabular}[l]{@{}l@{}}NAS-Bench-201\\Top Architecture \cite{Dong2020NAS-Bench-201}\end{tabular}         & 5.63 & 0.16 $\pm$ 0.13  & {-}    & 1.1     & 26.49 & 0.18 $\pm$ 0.13   & {-}    & 1.3  & 52.69 & 0.08 $\pm$ 0.12 & {-}                & 0.9    \\
 \midrule
RS-L                          & 10.73&1.06 $\pm$ 0.10  & 0.27 & 34.4  & 31.95 & 0.28 $\pm$ 0.09 & 0.01 & 0.7  & 82.08   & 0.01 $\pm$ 0.01 & 0.26 & 0.5     \\
RS-M                          & 21.67&0.09 $\pm$ 0.01& 0.01 & 0.3   & 46.82 & 0.06 $\pm$ 0.01 & 0.01 & 0.8  & 60.02     & 0.03 $\pm$ 0.01 & 0.01 & 1.3     \\ 
RS                        & 7.96 &0.03 $\pm$ 0.01& 0.01& 0.6   & 47.33 & 0.01 $\pm$ 0.01 & 0.01& 0.4  & 64.33       & 0.02 $\pm$ 0.01 & 0.01 & 0.7      \\
\midrule
Ours ($\lambda$=0)            & 6.05 &2.08 $\pm$ 0.04 & 1    & 43.9  & 29.44 & 1.67 $\pm$ 0.02   & 0.67 & 49.5 & 58.45  & 0.75 $\pm$ 0.01 & 1.29 & 224.7      \\
Ours ($\lambda$=0.25)         & 5.49 &0.97 $\pm$ 0.02 & 1.23 & 32.5 & 28.27 & 0.92 $\pm$ 0.35  & 1.9  & 60.4 & 57.33   & 3.43 $\pm$ 0.04 & 3    & 171.1      \\
Ours ($\lambda$=0.50)         & 4.64 &0.26 $\pm$ 0.04 & 1.36 & 13.9  & ${25.45}$ & 1.81 $\pm$ 0.02  & 2.7  & 51.5 & 49.12   & 0.19 $\pm$ 0.01 & 5.01 & 12.8       \\
Ours ($\lambda$=0.75)         & ${4.23}$ &0.33 $\pm$ 0.01 & 1.03 & 6.7  & 25.99 & 0.39 $\pm$ 0.18  & 2.8  & 32.3 & ${45.78}$   & 0.90 $\pm$ 0.01 & 4.79 & 42.7      \\
Ours ($\lambda$=1)            & 4.81 &0.15 $\pm$ 0.01 & 0.1  & 2.5  & 26.52 & 0.16 $\pm$ 0.01  & 0.01 & 2.2  & 51.17      & 0.76 $\pm$ 0.01 & 0.12 & 31.5      \\ \midrule
Ours (best)+AA$\dagger$    & $\mathbf{2.96}$ &0.33 $\pm$ 0.01 & 1.03 & 6.7 & $\mathbf{20.94}$  & 1.81 $\pm$ 0.02  & 2.7  & 51.5  & $\mathbf{43.35}$  & 0.90 $\pm$ 0.01 & 4.79 & 42.7   \\ \bottomrule
\end{tabular}%
}\\[2pt]
\raggedright \tiny $\dagger$~Results obtained by training the best model found in each data set: $\lambda=0.75$ for CIFAR-10 and ImageNet16-120 and $\lambda=0.5$ for CIFAR-100 with AutoAugment for 1500 epochs.

\end{table*}

\subsection{Final Training}
\label{subsec:training}
To evaluate the final architecture, we follow common training procedures \cite{liu2018darts,ENAS,real2019regularized}. However, we do not take advantage of drop path and other well-engineered training protocols that hide the contributions of the search strategy and the search space \cite{Yang2020NAS}, or that require forcing architectures to have specific schemes, e.g., auxiliary towers \cite{DBLP:conf/iclr/LarssonMS17}, as suggested by NAS best practices in order to showcase the true contributions of the proposed method \cite{DBLP:journals/corr/abs-1909-02453,DBLP:conf/uai/LiT19,Yang2020NAS,wan2022on}. 

Final architecture is trained for 600 epochs with batch size $96$. We use SGD optimization, with an initial learning rate, $\eta=0.025$ annealed down to zero following a cosine schedule without restart \cite{DBLP:conf/iclr/LoshchilovH17}, momentum of $0.9$ \cite{DBLP:conf/icml/SutskeverMDH13}, weight decay of $3\times10^{-4}$ and cutout \cite{devries2017improved}.

\subsection{Search Space}
To create the search space, we use all the 34 models present in TorchVision version 0.8 \cite{torchvision}: AlexNet \cite{NIPS2012_c399862d}, DenseNet\{121,161,169,201\} \cite{huang2017densely}, GoogLeNet \cite{szegedy2015going}, MNASNet\{0.5,0.75,1,1.3\} \cite{tan2019mnasnet}, MobileNetv2 \cite{sandler2018mobilenetv2}, ResNet\{18,34,50,101,152\} \cite{he2016deep}, ResNext\{50,101\} \cite{xie2017aggregated}, ShuffleNetv2\{0.5,1.0,1.5,2.0\} \cite{ma2018shufflenet}, SqueezeNet\{1.0,2.0\} \cite{iandola2016squeezenet}, VGG\{11,11BN,13,13BN,16,16BN,19,19BN\} \cite{Simonyan15} and Wide ResNet\{50, 101\} \cite{Zagoruyko_2016}. For each model, a WDG is generated and associated fitness is calculated using the performance estimation method.

The resulting search space is extremely large, composed of 17 possible operations: \{$1\times1$, $3\times3$, $5\times5$, $7\times7$, $11\times11$\} convolution; \{$2\times2$, $3\times3$\} max pooling; $2\times2$ average pooling; \{$1\times1$, $6\times6$, $7\times7$\} adaptive average pooling, $\{4096, 1024, 1000\}$ linear, $\{0.2, 0,5\}$ dropout and skip-connection. In the WDGs, $\mathbf{G(Q,E,H)}$, the inner-state choices for a given node, e.g., the output channels for a convolutional layer, are also present.

\sloppy As we do not bound the search either by defining architecture's outer-skeletons or the number of layers, the search space is extremelly complex, meaning that a generated architecture can be of any length and scheme. More, in our experiments, we create a search space for each evaluated data set, as the CNNs yield different fitnesses for different problems. Also, by using information about the probabilities of state-transition and the inner-states, LCMNAS leverages existing information to efficiently guide the search and to design both the architecture and the layer's parameters without requiring heuristics. Future works can use this search space without leveraging such information and solely focus on the possible operations to design cell-based architectures.

\subsection{Results and Discussion}
To evaluate the proposed method, we conducted extensive experiments using a single 1080Ti GPU. First, following the best practices proposed, we evaluate the performance of Random Search (RS) to infer the complexity of the search space \cite{DBLP:conf/uai/LiT19}. For this, we evaluate RS by randomly sampling architectures from the search space, $\mathcal{A}$, and two RSs based on the proposed evolutionary strategy: RS-L, that randomly samples layer components, and RS-M, that randomly samples models to serve as parents. The results for these three methods are shown in the second block of Table \ref{tab:totalresults}. As the three RSs attain high values for the test error, it shows that the search space is complex, and unlike other search spaces, RS is not sufficient to design competitive architectures \cite{DBLP:conf/uai/LiT19,Yang2020NAS}. Then, we evaluated the proposed method, by giving different importance's ($\lambda$) to the two components of the mixed-performance estimation. Results are presented in Table \ref{tab:totalresults}, third block. Following the best practices suggested in \cite{Yang2020NAS}, we first present the test errors without any added training protocol. From these, we see that a combination of both evaluating an architecture based on its trainability using low-fidelity estimates, and its modelling capabilities at initialization stage, yields the best results. Moreover, higher $\lambda$ values serve as regularization, reducing the number of parameters in the generated architectures. The search cost is lower when $\lambda=1$, due to the fast evaluation of untrained networks. Differences in search cost between data sets are due to the unconstrained properties of the search. As for noisier data sets, larger models tend to be generated, thus taking more time to evaluate. Also, the larger size of ImageNet16-120 also slows the evaluation. For the best architecture found in each dataset ($\lambda=0.75$ in CIFAR-10 and ImageNet16-120 and $\lambda=0.5$ in CIFAR-100), we also show the test error by training the architecture for 1500 epochs with AutoAugment \cite{cubuk2018autoaugment}, attaining 2.96\% test error in CIFAR-10, 20.94\% in CIFAR-100 and 43.35\% in ImageNet16-120. By comparing the proposed method with ResNet, and the best possible architecture in the NAS-Bench-201 (first block in Table \ref{tab:totalresults}), it is possible to see that the proposed method heavily out-performs existing state-of-the-art manually designed CNNs and all cell-based CNNs present in NAS-Bench-201. Moreover, comparing with RS is indicative of the effectiveness of the search, where the lower test errors show that LCMNAS effectively searches through the complex space of architectures.

\begin{table}[!t]
\centering
\caption{Comparison of different methods on CIFAR-10. First block present a state-of-the-art human-designed CNN. Second block presents the results of proposals that perform macro-search. Third block presents the proposed method. For each method, the test error in percentages, the search cost in terms of GPU days and the size of the architecture in millions of parameters is shown. Cells with a slash mean that the categorization does not apply.\label{tab:cifar10comparison}}
\resizebox{1\columnwidth}{!}{%
\begin{tabular}{lcccc} 
\toprule
\textbf{Method} & \begin{tabular}[r]{@{}c@{}}\textbf{Test Error}\\ (\%) $\downarrow$\end{tabular} & \begin{tabular}[c]{@{}c@{}}\textbf{Search Cost}\\ (GPU Days) $\downarrow$\end{tabular} & \begin{tabular}[c]{@{}c@{}}\textbf{Params}\\ (M) $\downarrow$\end{tabular} & \begin{tabular}[c]{@{}c@{}}\textbf{Search}\\ \textbf{Method}\end{tabular} \\
\midrule
ResNet \cite{he2016deep} & $6.43$ & {-} & $1.7$ & manual \\
\midrule
ConvFabrics \cite{saxena2016convolutional} & $7.43$ & {-} & $21.2$ & {-} \\
MetaQNN \cite{baker2016designing} &  $6.92$ & $100$ & $11.2$ & RL \\
NAS \cite{DBLP:journals/corr/ZophL16} & $4.47$ & $22400$ & $7.1$ & RL \\
NAS + more filters \cite{DBLP:journals/corr/ZophL16} & $3.65$ & $22400$ & $37.4$ & RL \\ 
Large-scale Evolution \cite{real2017large} & $5.40$ & $2750$ & $5.4$ & EA\\ 
SMASH \cite{brock2017smash} & $4.03$ & $1.5$ & $16.0$ & OS \\
Net Transformation \cite{cai2018efficient} & $5.70$ & $10$ & $19.7$ &  RL \\
Super Nets \cite{veniat2018learning} & $9.21$ & {-} & {-} & {-} \\
ENAS \cite{ENAS} & $4.23$ & $0.32$ & $21.3$ & RL \\
ENAS + more channels \cite{ENAS} & $3.87$ & $0.32$ & $38.0$ & RL \\ 
EPNAS \cite{perez2018efficient} & $5.14$ & $1.2$ & $5.9$ & SMBO \\
NSGA-NET \cite{DBLP:conf/gecco/LuWBDDGB19} & $3.85$ & $8$ & $3.3$ & EA \\
RandGrow  \cite{humacro} & $2.93$ & $6$ & $3.1$ & RS \\
LEMONADE \cite{elsken2018efficient} & $3.60$ & $80$ & $8.9$ & EA \\
Petridish \cite{hu2019efficient} & $\mathbf{2.83}$ & $5$ & $2.2$ & GB \\
\midrule
\textbf{Ours} $(\lambda=0.75)$+AA & ${2.96}$ & $1.03$ & $6.7$ & EA 
\\ \bottomrule
\end{tabular}
}
\end{table}

\begin{table}[!tbh]
\centering
\caption{Test error (\%) obtained by transferring the best architectures found in one data set to other data sets without any modifications. Architectures were trained on the new data set from scratch. \label{tab:transferring}}
\resizebox{1\columnwidth}{!}{%
\setlength\extrarowheight{-0pt}
\begin{tabular}{lccc}
\toprule
\multicolumn{1}{c}{\multirow{2}{*}{{Searched On}}} & \multicolumn{3}{c}{Transferred To}                                             \\ \cmidrule{2-4}
                                   & CIFAR-10 & CIFAR-100 & ImageNet16-120 \\ \midrule
CIFAR-10                                & $\mathbf{4.23}$& $\mathbf{24.42}$     & $53.45$      \\
CIFAR-100                               & $5.09$         & $25.45$     & $55.42$      \\
ImageNet16-120                          & $8.15$         & $30.39$     & $\mathbf{45.78}$      \\ \bottomrule                     
\end{tabular}
}
\end{table}

Table \ref{tab:cifar10comparison} compares the proposed method with different NAS methods that perform macro-search using CIFAR-10. It is important to note that most of these methods heavily rely on tuned search spaces and constraints, such as architecture skeletons, number of layers and forced initial and final operations, which has shown to hide the real contributions of the search strategy \cite{Yang2020NAS}. From this table, it is possible to see that LCMNAS achieves high performances (low test error) while at the same time being orders of magnitude faster (search cost). LCMNAS closely relates with large-scale evolution \cite{real2017large}, since both apply evolutionary strategies and heavily reduce the amount of human intervention in the decision process. By direct comparison, LCMNAS achieved a lower test error by 2.44\%, while being orders of magnitude faster (2669x faster). By directly comparing with RandGrow and Petridish, the closest in terms of test error, we see that LCMNAS is 5x faster, while achieving similar test errors. However, RandGrow \cite{humacro} and Petridish \cite{hu2019efficient} rely on DropPath, which in RandGrow's case, reduced its' base test error from 3.38\% to 2.93\%. Regarding CIFAR-100, few proposals directly search on it, instead transfer architectures from CIFAR-10. As comparison, Large-scale evolution \cite{real2017large} achieved a test error of 23.7\%, which is 2.76\% higher than the 20.94\% obtain by LCMNAS.

As most NAS methods focus on designing architectures in one data set and then transferring it to larger ones, we also study this using the best-generated architectures, without AutoAugment. However, contrary to common practices, we do not change the architectures when transferring them, to emphasize the true contributions of the search strategy. What we found is that for similar data sets, i.e., CIFAR-10 and CIFAR-100, the architecture found in the first and transferred to the second, out-performs the architecture found directly searching on the second. We justify this due to the similar nature of the data sets, whereas CIFAR-100 is a more demanding data set, thus searching on that data set is harder. However, the same does not apply when transferring CIFAR-10 or CIFAR-100 to ImageNet16-120 and the other way around. The result of such transference falls heavily short when compared to directly searching in the desired data set. These results are shown in Table \ref{tab:transferring}. From these, we conclude that directly searching is better than transferring if there are no similar and simpler data sets available that might aid in the search process.

Regarding the architecture's design, a visualization of the best ones can be seen in the Appendix. LCMNAS found that placing batch normalization layers after convolutional layers was an important building block, and usually ReLu also follows, meaning that after a convolutional layer, regularization layers are important. In CIFAR-10, LCMNAS specifically designed a very deep network with a small number of output channels, thus controlling the number of parameters. In the middle of this architecture, a dropout layer and a linear layer were added. We hypothesize that this served the purpose of regularization by not only dropping connections but also creating linear representations of the feature maps. In many architectures, a pattern of having convolutional layers with larger kernel sizes ($7\times7$) interspersed with smaller ones ($1\times1$ and $3\times3$) is seen. Also common to all architectures, LCMNAS tends to add dropout layers and reduction mechanisms, e.g., adaptive average pooling, at the end of the architecture, especially in ImageNet16-120, the noisier data set.



\begin{table}[!tbh]
\centering
\setlength\extrarowheight{-2pt}
\caption{Test error (\%) of the best architectures found on CIFAR-10 using different search epochs $e$ and fixed $\lambda=0.75$.\label{tab:searchepochs}}
\begin{tabular}{lccc}
\toprule
\multirow{2}{*}{\begin{tabular}[c]{@{}c@{}}Search Epochs\\$e$\end{tabular}} & \multicolumn{2}{c}{CIFAR-10}  \\ \cmidrule(lr){2-3}
                                   & Test Error (\%) $\downarrow$& Params (M) $\downarrow$ \\ \midrule 
$e = 1$                            & $6.33$         & $21.4$      \\
$e = 2$                            & $4.92$         & $170.1$      \\
$e = 3$                            & $4.69$         & $44.1$      \\
$e = 4$                            & $4.23$         & $6.7$      \\ \bottomrule                      
\end{tabular}
\end{table}

\subsection{Ablation Studies}

We extend the studies of LCMNAS by evaluating the importance of the epochs used to partially train a generated architecture, using $\lambda=0.75$ (results shown in Table \ref{tab:searchepochs}). Naturally, reducing $e$ implied an increase in the final test error and in the number of parameters of the final architecture. This can be justified by the difficulty in measuring the trainability of the architectures with low training epochs. Furthermore, in Table \ref{tab:genparams}, we evaluated the impact of the number of generations, $g$, and population per generation, $p$, using $\lambda=1$. The resulting test error shows that $g=50$ and $p=100$ yield the best results and that these parameters can influence the final result by a significant amount. In a perfect scenario, where computational costs are not considered, one could use a large $e$ and evaluate more architectures by increasing $g$ and $p$.

Also, we evaluate the diversity of generated architectures by performing ensembling \cite{DBLP:conf/nips/ZaidiZEHHT21}. For this, we ensemble the architectures found, using different $\lambda$ values (Table \ref{tab:ensemble}) as indication of the architectures' diversity \cite{zaidi20}. Ensemble sizes $k$, go from the top-1 architecture ($k=1$) to the top-5 architecture found. The resulting test error was obtained using weighted majority voting, with the accuracies attained while training used as weights. Results show that for all data sets evaluated, incrementally adding lower-performant architectures yields better results than the top-1 architecture alone. This experiment validates the premise that less constrained macro-search has benefits due to its' inherent architecture generation diversity.

\begin{table}[!tbh]
\centering
\caption{Test error (\%) in CIFAR-10 using $\lambda=1$ for different number of generations, $g$, and population sizes, $p$.\label{tab:genparams}}
\resizebox{1\columnwidth}{!}{ %
\setlength\extrarowheight{-3.5pt}
\begin{tabular}{cccc} 
\toprule
\multicolumn{2}{c}{Parameters} & \multicolumn{2}{c}{CIFAR-10} \\
\cmidrule(lr){1-2} \cmidrule(lr){3-4}
Generations ($g$) & Population ($p$) & \multicolumn{1}{c}{Test Error (\%) $\downarrow$}  & \multicolumn{1}{c}{Params (M) $\downarrow$} \\ \midrule
$5$              & $5$             & $16.52$           & $0.3$       \\ \midrule
$10$             & $10$            & $5.46$            & $3.9$       \\ 
$10$             & $25$            & $5.25$            & $3.6$       \\ 
$10$             & $50$            & $6.97$            & $0.9$       \\
$10$             & $100$           & $6.83$            & $12.7$      \\ \midrule
$15$             & $15$            & $6.08$            & $0.4$       \\
$15$             & $25$            & $5.58$            & $8.7$       \\
$15$             & $50$            & $6.16$            & $7.6$     \\
$15$             & $100$           & $5.37$            & $13.9$  \\ \midrule 
$25$             & $25$            & $5.49$            & $2$         \\
$25$             & $50$            & $6.52$            & $3.1$       \\
$25$             & $100$           & $4.97$            & $8.4$       \\ \midrule 
$50$             & $50$            & $5.24$            & $4.7$       \\
$50$             & $100$           & $4.81$            & $2.5$      \\
\midrule
$100$            & $50$            & $6.69$            & $5.3$       \\
$100$            & $100$           & $5.94$            & $13.6$      \\
$100$            & $250$           & $5.54$            & $55.9$      \\
\midrule
$150$            & $150$           & $6.19$            & $3.3$      \\
$150$            & $250$           & $6.46$            & $2.2$       \\
\midrule
$250$            & $100$           & $5.27$            & $36.1$      \\ 
$250$            & $250$           & $6.14$            & $4.8$       \\
\bottomrule
\end{tabular}
}
\end{table}

\begin{table}[!tbh]
\caption{Ensemble test error (\%) for CIFAR-10, CIFAR-100 and ImageNet-16-120 with different ensemble sizes $k$. Architectures used to perform the ensemble are the final ones found by searching with different $\lambda \in \{0,0.25,0.5,0.75,1\}$.\label{tab:ensemble}}
\centering
\resizebox{1\columnwidth}{!}{%
\begin{tabular}{lccc}
\toprule
\multirow{2}{*}{\parbox{1.2cm}{Ensemble Size ($k$)}} & \multicolumn{3}{c}{Test Error $(\%)$ $\downarrow$}                                             \\ \cmidrule{2-4}
                                   & CIFAR-10 & CIFAR-100 & ImageNet16-120 \\ \midrule
$k=1$                                & $4.23$         & $25.45$     & $45.78$      \\
$k=2$                                & $4.23$         & $25.45$     & $45.78$      \\
$k=3$                                & $3.67$         & $22.64$     & $\mathbf{44.67}$      \\
$k=4$                                & $\mathbf{3.51}$         & $22.01$     & $44.85$      \\
$k=5$                                & $3.59$& $\mathbf{21.52}$& ${44.72}$       \\ \bottomrule                     
\end{tabular}
}
\end{table}

\section{Designing Cells: Experiments and Results Analysis}
To further allow comparison with the state-of-the-art, we evaluate the effectiveness of the proposed NAS algorithm in the problem of cell-based NAS. For this, we utilise two different search spaces: NAS-Bench-201 \cite{Dong2020NAS-Bench-201} and TransNAS-Bench-101 \cite{DBLP:conf/cvpr/DuanCXCLZL21} benchmarks. These benchmarks were designed to have tractable NAS search spaces with metadata for the training of thousands of architectures within those search spaces.

\subsection{Search Spaces}
\textbf{NAS-Bench-201} fixes the search space as a cell-based design with 5 possible operations: zeroize, skip connection, $1\times1$ and $3\times3$ convolution, and $3\times3$ average pooling layer. The cell design comprises six edges and four nodes, where an edge represents a possible operation through two nodes. By fixing the cell size and the operation pool, the search space comprises $5^6 = 15625$ possible cells. To form entire networks, the cells are replicated in an outer-defined skeleton. NAS-Bench-201 provides information regarding the training and performance of all possible networks in the search space in three different data sets: CIFAR-10, CIFAR-100 and ImageNet16-120, thus proposing a controlled setting that allows different NAS methods to be fairly compared, as they are forced to use the search space, training procedures and hyper-parameters.

\textbf{TransNAS-Bench-101} is a benchmark that provides architecture's performances across seven vision tasks including classification, regression, pixel-level prediction and self-supervised tasks. The 7 tasks of this benchmark are: object classification, scene classification, autoenconding, surface normal, semantic segmentation, room layout and jigsaw. By having multiple tasks that are queryable with the same input, this benchmark provides the opportunity to evaluate NAS transferability between different tasks.
There are two types of search space in this benchmark, i.e., the widely-studied cell-based search space containing 4096 architectures and macro skeleton search space based on residual blocks containing 3256 architectures. Possible operations are: zeroize, skip connection, $1\times1$ and $3\times3$ convolution. Transnas-Bench-101 provides information regarding the training and performance of all possible networks in the search space using the same training protocols and hyper-parameters within each task.

\subsection{Results and Discussion}
For LCMNAS to perform micro-search in different search spaces, the search method was slightly modified to accommodate the forced rules that micro-search carry. First, the initial population is randomly sampled from the search space. Secondly, the search method samples a specific number of layers $l$, which for both NAS-Bench-201 and TransNas-Bench-101 is $l=6$. Finally, the search method focus solely on sampling layers (operations) instead of sampling both layers and associated parameters.

To evaluate LCMNAS on NAS-Bench-201, we set $P=10$ and $G=35$ to allow a fair comparison with other non-weight sharing NAS methods in terms of search time and to share similar settings as other evolutionary methods \cite{real2019regularized}. The results for searching on all NAS-Bench-201 datasets: CIFAR-10, CIFAR-100 and ImageNet16-120 are presented in Table \ref{table:benchmarkingnasbench201lcmnas}. Notably, LCMAS achieved competitive results, outperforming current state-of-the-art cell-based methods \cite{real2019regularized,liu2018darts}, while requiring less computation (search time) when compared to both non-weight sharing and weight sharing methods. Compared with GDAS \cite{DBLP:conf/cvpr/DongY19}, the current best weight sharing method in NAS-Bench-201 in terms of accuracy, LCMNAS outperforms GDAS by 0.54\%, 1.4\% and 3.77\% in CIFAR-10, CIFAR-100 and ImageNet16-120 respectively. The gains in terms of accuracy are more notable in noiser data sets, being ImageNet16-120 the noiser, where images are small, with a size of $16 \times 16$ pixels, with a training set of 140 thousand images spanning across 120 classes. More, LCMNAS is also more efficient than GDAS in terms of computation, requiring less than 40\% of the computation in terms of search time. When comparing non-weight sharing methods, LCMNAS outperforms REA \cite{real2019regularized} in all data sets, while also requiring less computation. More, LCMNAS is highly precise in finding good architectures, which is shown by small standard deviations in all data sets, especially in ImageNet16-120.

To further assess LCMNAS on a micro-search setting, we also evaluate its performance on TransNas-Bench-101 in all 7 different tasks. This evaluation contributes to validating its generability and transferability across different problems, which is a problem where NAS methods commonly fail \cite{Yang2020NAS,Dong2020NAS-Bench-201,wan2022on}. For this, we conducted two different experiments: i) directly searching on each task independently, and ii) performing transfer search. For the latter, we followed common procedures \cite{DBLP:conf/cvpr/DuanCXCLZL21}, where first the method searches on jigsaw and uses the final population as initialization for the evolution when searching on the other tasks. The results for both experiments are shown in Table \ref{table:comparisontransnas101lcmnas}. In the first block of the table results are shown for directly searching in each data set independently. In this, random search (RS) \cite{DBLP:journals/jmlr/BergstraB12} serves as baseline to whether NAS methods are actually generating good architectures and learning anything. The results obtained by LCMNAS show that is not only learning, but that it is capable of generating architectures that achieve the overall global best performance in all data sets. The results are also comparable with existing methods, where LCMNAS outperforms existing state-of-the-art results in all data sets, where improvements are significant, corresponding to an improvement of 3.6\% in the best cases when compared to Arch-Graph-Single \cite{DBLP:journals/corr/abs-2204-05941}.

For the task of transferability between data sets, results are shown in the second block of Table \ref{table:comparisontransnas101lcmnas}. Results show that that the cell-based search of LCMNAS is indeed transferable between different data sets, and capable of achieving state-of-the-art results. By searching on Jigsaw and then transfering to other data sets, LCMNAS was capable of generating cells that achieve the global best result in TransNas-Bench-101. More, LCMNAS outperforms other NAS methods in all data sets, where in some cases the performance gains when compared to the current state-of-the-art, Arch-Graph \cite{DBLP:journals/corr/abs-2204-05941}, are 2.3\%. The overall results in TransNas-Bench-101 support LCMNAS in terms of performance, generalibity and transferability, showing that it can efficiently be used to directly search or be transferred to new and different problems.


The results in all 10 data sets across the 2 benchmarks used to evaluate LCMNAS cell-search show that LCMNAS is an efficient approach to generating neural networks. In terms of performance, the generated architectures achieve state-of-the-art results in all data sets, while requiring less computation (time) to conduct the search.

\begin{table*}[t]
	\caption{Comparison of manually designed networks and several search methods evaluated using the NAS-Bench-201 benchmark. Performance is shown in terms of accuracy (\%) with mean$\pm$std, on CIFAR-10, CIFAR-100 and ImageNet-16-120. Search times are the mean time required to search for cells in CIFAR-10. Search time includes the time taken to train networks as part of the process where applicable. Table adapted from \cite{Dong2020NAS-Bench-201,lopes2021epenas,mellor2020neural}.\label{table:benchmarkingnasbench201lcmnas}}
	\footnotesize
	\setlength{\arrayrulewidth}{2pt}

	\resizebox{\textwidth}{!}{%
		\begin{tabular}{@{}lrllcllcll@{}} \hline  
			\multirow{2}{*}{Method} & \multirow{2}{*}{\shortstack{Search \\Time (s)}$\downarrow$}  & \multicolumn{2}{c}{CIFAR-10} & \phantom{} & \multicolumn{2}{c}{CIFAR-100} & \phantom{} & \multicolumn{2}{c}{ImageNet-16-120} \\
			\cmidrule{3-4} \cmidrule{6-7} \cmidrule{9-10}
			& & \multicolumn{1}{c}{Val. Acc (\%)$\uparrow$} & \multicolumn{1}{c}{Test Acc. (\%)$\uparrow$}  && \multicolumn{1}{c}{Val. Acc (\%)$\uparrow$} & \multicolumn{1}{c}{Test Acc. (\%)$\uparrow$} && \multicolumn{1}{c}{Val. Acc (\%)$\uparrow$} & \multicolumn{1}{c}{Test Acc. (\%)$\uparrow$} \\
			\midrule

            \multicolumn{10}{c}{\textbf{Manually designed}}\\
			ResNet  \cite{he2016deep}      & -  & \multicolumn{1}{c}{90.83} & \multicolumn{1}{c}{93.97} && \multicolumn{1}{c}{70.42} & \multicolumn{1}{c}{70.86} && \multicolumn{1}{c}{44.53} & \multicolumn{1}{c}{43.63} \\
            \midrule \midrule
			\multicolumn{10}{c}{\textbf{Weight sharing}}\\
			RSPS   \cite{DBLP:conf/uai/LiT19}     & 7587  & 84.16$\pm$1.69 & 87.66$\pm$1.69 && 59.00$\pm$4.60 & 58.33$\pm$4.34 && 31.56$\pm$3.28 & 31.14$\pm$3.88 \\
			DARTS-V1  \cite{liu2018darts}  & 10890 & 39.77$\pm$0.00 & 54.30$\pm$0.00 && 15.03$\pm$0.00 & 15.61$\pm$0.00 && 16.43$\pm$0.00 & 16.32$\pm$0.00 \\
			DARTS-V2  \cite{liu2018darts}  & 29902 & 39.77$\pm$0.00 & 54.30$\pm$0.00 && 15.03$\pm$0.00 & 15.61$\pm$0.00 && 16.43$\pm$0.00 & 16.32$\pm$0.00 \\
			GDAS     \cite{DBLP:conf/cvpr/DongY19}   & 28926 & 90.00$\pm$0.21 & 93.51$\pm$0.13 && 71.14$\pm$0.27 & 70.61$\pm$0.26 && 41.70$\pm$1.26 & 41.84$\pm$0.90 \\
			SETN      \cite{DBLP:conf/iccv/Dong019a}  & 31010 & 82.25$\pm$5.17 & 86.19$\pm$4.63 && 56.86$\pm$7.59 & 56.87$\pm$7.77 && 32.54$\pm$3.63 & 31.90$\pm$4.07 \\
			ENAS     \cite{ENAS}   & 13315 & 39.77$\pm$0.00 & 54.30$\pm$0.00 && 15.03$\pm$0.00 & 15.61$\pm$0.00 && 16.43$\pm$0.00 & 16.32$\pm$0.00 \\

			\midrule \midrule
			\multicolumn{10}{c}{\textbf{Non-weight sharing}}\\
			RS      \cite{DBLP:journals/jmlr/BergstraB12}  &  12000 & 90.93$\pm$0.36 & 93.70$\pm$0.36 && 70.93$\pm$1.09 & 71.04$\pm$1.07 && 44.45$\pm$1.10 & 44.57$\pm$1.25 \\
			REINFORCE \cite{DBLP:journals/ml/Williams92} &  12000 & 91.09$\pm$0.37 & 93.85$\pm$0.37 && 71.61$\pm$1.12 & 71.71$\pm$1.09 && 45.05$\pm$1.02 & 45.24$\pm$1.18 \\
			BOHB     \cite{DBLP:conf/icml/FalknerKH18} &  12000 & 90.82$\pm$0.53 & 93.61$\pm$0.52 && 70.74$\pm$1.29 & 70.85$\pm$1.28 && 44.26$\pm$1.36 & 44.42$\pm$1.49 \\
			REA      \cite{real2019regularized} & 12000 &91.19$\pm$0.31  &93.92$\pm$0.30  &&71.81$\pm$1.12  & 71.84$\pm$0.99 &&45.15$\pm$0.92 & 45.54$\pm$1.03

			\\
			\textbf{LCMNAS (ours)}$\dagger$ & 11521 &\textbf{91.22}$\pm$\textbf{0.17}  &\textbf{94.05}$\pm$\textbf{0.07}  &&\textbf{71.96}$\pm$\textbf{0.96}  & \textbf{72.01}$\pm$\textbf{0.82} && \textbf{44.55}$\pm$\textbf{0.78}& \textbf{45.61}$\pm$\textbf{0.08}
			\\
			\bottomrule
		\end{tabular}
	}
	\begin{flushleft}
	\footnotesize $\dagger$~Results of 10 runs using the same settings: $P/G=10/35$, using a single 1080Ti GPU.\\
	\end{flushleft}
\end{table*}

\begin{table*}[!t]
 \caption{\label{table:comparisontransnas101lcmnas}Performance comparison of different NAS methods on TransNAS-Bench-101. The first block shows the results for direct search. The second block shows the transferred versions of different methods, which are pretrained on the least time-consuming task, i.e., Jigsaw. The final row shows the possible best result (global best) in each task.}

\begin{centering}
\small
\resizebox{\textwidth}{!}{%
\begin{tabular}{c|l|cccccccc}
\toprule 
& \multicolumn{1}{c|}{{Tasks}} & {Cls. Object} & {Cls. Scene} & {Autoencoding} & {Surf. Normal}  & {Sem. Segment.} & {Room Layout} & {Jigsaw} \tabularnewline \midrule
& \multicolumn{1}{c|}{{Metric}} & \textit{Acc. (\%)} $\uparrow$ & \textit{Acc. (\%)} $\uparrow$ & \textit{SSIM} $\uparrow$ & \textit{SSIM} $\uparrow$ & \textit{mIoU} $\uparrow$ & \textit{L2 loss} $\downarrow$ & \textit{Acc. (\%)} $\uparrow$ \tabularnewline \midrule
& {RS \cite{DBLP:journals/jmlr/BergstraB12}} & {45.16$\pm$0.4} & {54.41$\pm$0.3} & {55.94$\pm$0.8} & {56.85$\pm$0.6} & {25.21$\pm$0.4} & {61.48$\pm$0.8} & {94.47$\pm$0.3} \tabularnewline
& {REA \cite{real2019regularized}} & {45.39$\pm$0.2} & {54.62$\pm$0.2} & {56.96$\pm$0.1} & {57.22$\pm$0.3} & {25.52$\pm$0.3} & {61.75$\pm$0.8} & {94.62$\pm$0.3} \tabularnewline
\parbox[t]{2mm}{\multirow{5}{*}{\rotatebox[origin=c]{90}{Direct Search}}} & {PPO \cite{DBLP:journals/corr/SchulmanWDRK17}} & {45.19$\pm$0.3} & {54.37$\pm$0.2} & {55.83$\pm$0.7} & {56.90$\pm$0.6} & {25.24$\pm$0.3} & {61.38$\pm$0.7} & {94.46$\pm$0.3}\tabularnewline
& {DT} & {\ 42.03$\pm$5.0 } & {\ 49.80$\pm$8.6 } & {\ 51.20$\pm$3.3 } & {\ 55.03$\pm$2.7 } & {\ 22.45$\pm$3.2 } & {\ 66.98$\pm$2.3 } & {\ 88.95$\pm$9.1 } \tabularnewline
& {BONAS \cite{DBLP:conf/nips/ShiPXLKZ20}$\dag$} & {45.50} & {54.56} & {56.73} & {57.46} & {25.32} & {61.10} & {94.81} \tabularnewline
& {weakNAS \cite{DBLP:conf/nips/WuDCCLYWLCY21}$\dag$} & {45.66} & {54.72} & {56.77} & {57.21} & {25.90} & {60.31} & {94.63} \tabularnewline
&{Arch-Graph-single \cite{DBLP:journals/corr/abs-2204-05941}$\dag$} & {45.48} & {54.70} & {56.52} & {57.53} & {25.71} & {61.05} & {94.66} \tabularnewline
& {LCMNAS (Ours)$\dag$} & \textbf{{46.32}} & \textbf{{54.94}} & \textbf{{57.72}} & \textbf{{59.62}} & \textbf{{26.27}} & \textbf{{59.38}} & \textbf{{95.37}} \tabularnewline
\midrule


& {REA-t \cite{real2019regularized}} & {45.51$\pm$0.3} & {54.61$\pm$0.2} & {56.52$\pm$0.6} & {57.20$\pm$0.7} & {25.46$\pm$0.4} & {61.04$\pm$1.0} & {-} \tabularnewline
& {PPO-t \cite{DBLP:journals/corr/SchulmanWDRK17}} & {44.81$\pm$0.6} & {54.15$\pm$0.5} & {55.70$\pm$1.5} & {56.60$\pm$0.7} & {24.89$\pm$0.5} & {62.01$\pm$1.0} & {-} \tabularnewline
\parbox[t]{2mm}{\multirow{1}{*}{\rotatebox[origin=c]{90}{Transfer Search}}}& {CATCH \cite{DBLP:conf/eccv/ChenDCXCLZL20}} & {45.27$\pm$0.5} & {54.38$\pm$0.2} & {56.13$\pm$0.7} & {56.99$\pm$0.6} & {25.38$\pm$0.4} & {60.70$\pm$0.7} & {-} \tabularnewline
& {BONAS-t \cite{DBLP:conf/nips/ShiPXLKZ20}$\dag$} & {45.38} & {54.57} & {56.18} & {57.24} & {25.24} & {60.93} & {-} \tabularnewline
& {weakNAS-t \cite{DBLP:conf/nips/WuDCCLYWLCY21}$\dag$} & {45.29} & {54.78} & {56.90} & {57.19} & {25.41} & {60.70} & {-} \tabularnewline
& {Arch-Graph-zero \cite{DBLP:journals/corr/abs-2204-05941}$\dag$} & {45.64} & {54.80} & {56.61} & {57.90} & {25.73} & {60.21} & {-} \tabularnewline
& {Arch-Graph \cite{DBLP:journals/corr/abs-2204-05941}$\dag$} & {45.81} & {54.90} & {56.58} & {58.27} & {25.69} & {60.08} & {-} \tabularnewline
& {LCMNAS-t (Ours)$\dag$} & \textbf{{46.32}} & \textbf{{54.94}} & \textbf{{57.72}} & \textbf{{59.62}} & \textbf{{26.27}} & \textbf{{59.38}} & {-} \tabularnewline
\midrule
& {Global Best} & {46.32} & {54.94} & {57.72} & {59.62} & {26.27} & {59.38} & {95.37} \tabularnewline
\bottomrule  

\end{tabular}
}
\par\end{centering}
\begin{flushleft}
\footnotesize $\dag$~Results provided for the best run only.\\
\end{flushleft}
\end{table*}

\subsection{Ablation Studies}
Since LCMNAS relies on a hybrid performance estimation approach to decide which architectures are worth keeping, we further evaluate its value on cell-based search. For this, we use the NAS-Bench-201 benchmark and evaluate the Kendall's Tau correlation $\tau$ between between the proposed mixed-performance estimation and the final validation accuracy for the first 1000 architectures in each data set, using different $\lambda$ values and partial training epochs $e$. The results presented in Table \ref{tab:nasbenchtaucorrlcmnas} show the effectiveness of the performance estimation strategy. Notably, when $\lambda=0$, the performance estimation strategy relies solely on the partial training of architectures to evaluate them. In contrast, $\lambda=1$ relies only on scoring the architectures at initialization stage, which has higher $\tau$ correlations when compared to partial training with $e=4$. 

On all data sets it is clear that combining both components on the performance estimation strategy yields the best correlation, where $\lambda=0.75$ tends to yield the best $\tau$ correlation. This means that more importance is giving to scoring architectures at initialization stage than to the partial training. Expectedly, increasing the number of epochs $e$ contributes significantly to the final $\tau$ correlation, as using only one epoch of partial training does not add valuable information to the scoring. 

The results on Table \ref{tab:nasbenchtaucorrlcmnas} show that the combination of both partial training and scoring architectures at initialization stage is a good indicator of their final performance, thus validating the importance of the performance estimation mechanism.


\begin{table}[!t]
\caption{Kendall's Tau correlation ($\tau$) across each of NAS-Bench-201 datasets for different number of train epochs ($e$) and Lambda ($\lambda$) values. The results show that on all settings, the combination of both components of the performance estimation strategy leads to a higher correlation with regards to the final validation accuracy of the generated architectures.\label{tab:nasbenchtaucorrlcmnas}}
\resizebox{1\columnwidth}{!}{%
\begin{tabular}{llllll}
\toprule
\multirow{2}{*}{Lambda ($\lambda)$} & \multicolumn{5}{c}{Train epochs ($e$)} \\ \cmidrule{2-6}
 & 0 & 1 & 2 & 3 & 4 \\ \midrule
 \multicolumn{6}{c}{\textbf{CIFAR-10}}
 \\\midrule
0 & - & 0.298 & 0.392 & 0.458 & 0.530 \\
0.25 & - & 0.550 & 0.535 & 0.572 & 0.614 \\
0.5 & - & 0.581 & 0.567 & 0.598 & 0.629 \\
0.75 & - & 0.586 & 0.579 & 0.599 & 0.630 \\
1 & 0.574 & - & - & - & - 
\\ \midrule
 \multicolumn{6}{c}{\textbf{CIFAR-100}}\\ \midrule
0 & - & 0.045 & 0.254 & 0.326 & 0.395 \\
0.25 & - & 0.524 & 0.547 & 0.553 & 0.561 \\
0.5 & - & 0.544 & 0.560 & 0.565 & 0.571 \\
0.75 & - & 0.555 & 0.563 & 0.567 & 0.574 \\
1 & 0.558 & - & - & - & - 
\\ \midrule
 \multicolumn{6}{c}{\textbf{ImageNet16-120}}\\ \midrule
0 & - & 0.221 & 0.286 & 0.371 & 0.389 \\
0.25 & - & 0.526 & 0.540 & 0.552 & 0.559 \\
0.5 & - & 0.536 & 0.551 & 0.558 & 0.565 \\
0.75 & - & 0.545 & 0.555 & 0.562 & 0.566 \\
1 & 0.544 & - & - & - & - 
\\ \bottomrule
\end{tabular}
}
\end{table}


\section{Conclusion}

In this work, we propose a NAS approach that is capable of performing for cell-based search and unconstrained macro-search. For this, we design three novel components for the NAS method. For the search space design, we propose a method that autonomously generates complex search spaces by creating Weighted Directed Graphs with hidden properties from existing CNNs. The proposed search strategy is capable of performing both micro and macro-architecture search, via evolution, without requiring human-defined restrictions, such as outer-skeleton, initial architecture, or heuristics. To quickly evaluate generated architectures, we propose the use of a mixed-performance strategy that combines information about architectures at initialization stage with information about their validation accuracy after a partial train on a partial data set. Our experiments show that LCMNAS generates state-of-the-art architectures both in cell-based search, outperfoming current methods in 10 different datasets, and in macro-based search, where it is capable of generating architectures from scratch with minimal GPU computation that achieved test errors of 2.96\% in CIFAR-10, 20.94\% in CIFAR-100 and 43.35\% in ImageNet16-120. We also study the importance of different NAS components and draw insights from architecture design choices.

We hope that this work serves the purpose of pushing NAS boundaries to less constrained spaces, where human-expertize for the design of inner-architecture parameters and search spaces is reduced, while at the same time generating architectures in a very efficient way, thus allowing a step towards wide-spread use of NAS for different problems and data sets.



{\appendices
\section*{Visualization of the Best Architectures Found}
\label{sec:appendixlcmnasbest}
Figs. \ref{fig:best_cifar10}, \ref{fig:best_cifar100}, \ref{fig:best_imgnet16120}, show the architecture of the best models found by searching on CIFAR-10, CIFAR-100 and ImageNet16-120, respectively.
\begin{figure}[!tbh]
    \centering
    \includegraphics[height=18.5cm]{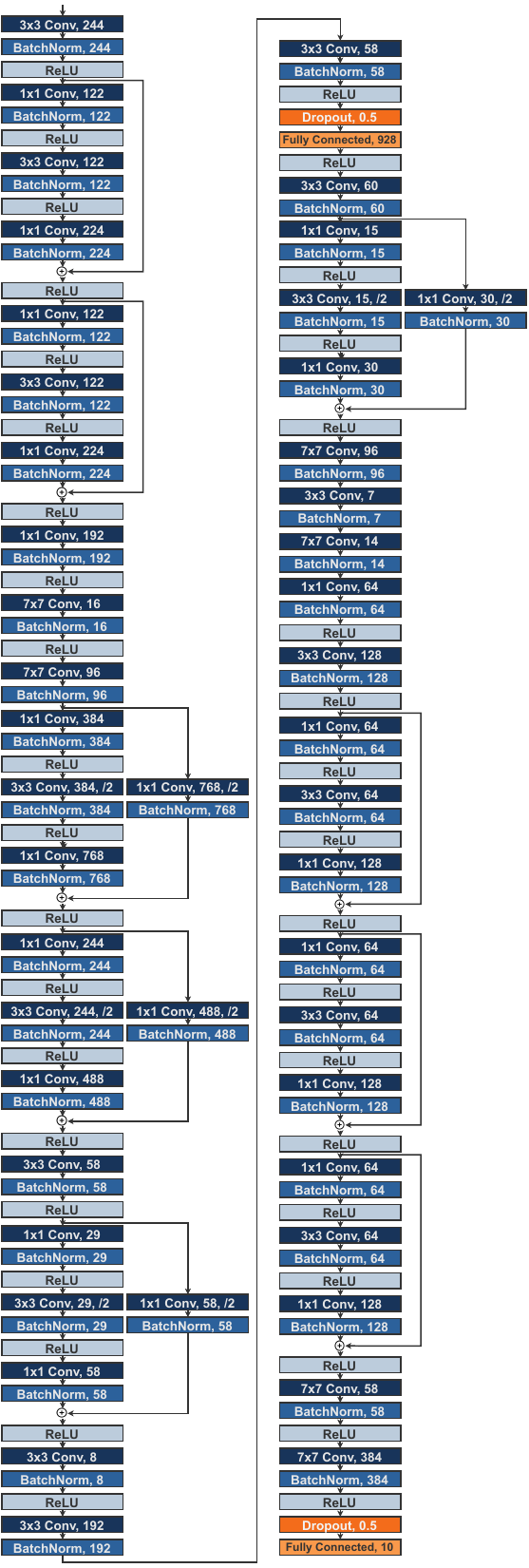}
    \caption{Architecture of the best model found in CIFAR-10 ($\lambda=0.75$).}
    \label{fig:best_cifar10}
\end{figure}

\begin{figure}[!tbh]
    \centering
    \includegraphics[]{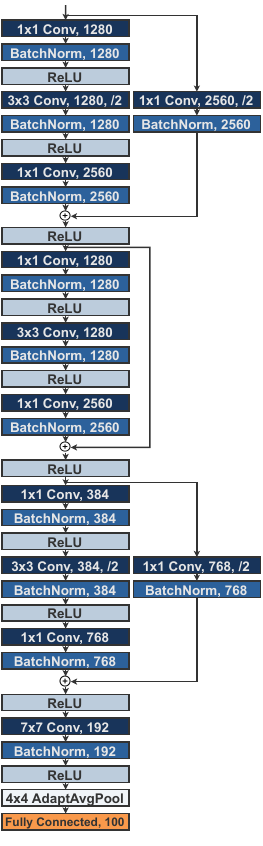}
    \caption{Architecture of the best model found in CIFAR-100 ($\lambda=0.50$).}
    \label{fig:best_cifar100}
\end{figure}

\begin{figure}[!tbh]
    \centering
    \includegraphics[height=18.5cm]{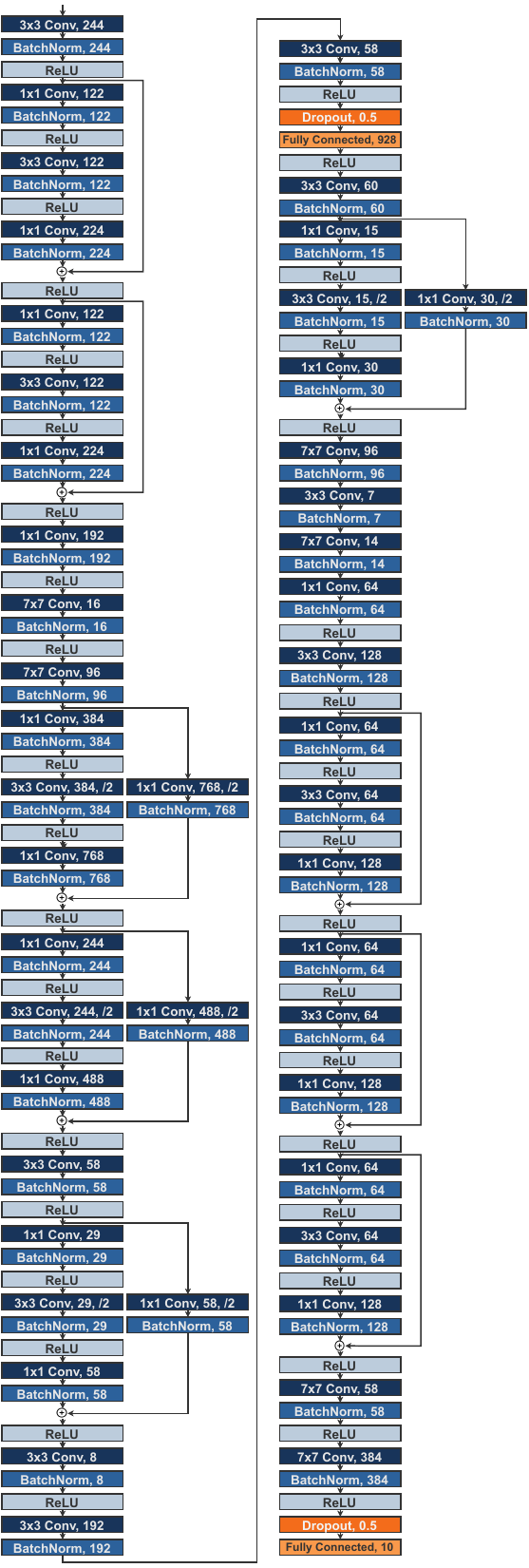}
    \caption{Architecture of the best model found in ImageNet16-120 ($\lambda=0.75$).}
    \label{fig:best_imgnet16120}
\end{figure}

}

{\small
\bibliographystyle{ieee_fullname}
\bibliography{bib}
}

\vfill

\end{document}